\documentclass{article}

\usepackage[preprint]{neurips_data_2023}

\usepackage[utf8]{inputenc} % allow utf-8 input
\usepackage[T1]{fontenc}    % use 8-bit T1 fonts
\usepackage{hyperref}       % hyperlinks
\usepackage{url}            % simple URL typesetting
\usepackage{booktabs}       % professional-quality tables
\usepackage{amsfonts}       % blackboard math symbols
\usepackage{nicefrac}       % compact symbols for 1/2, etc.
\usepackage{microtype}      % microtypography
\usepackage{xcolor}         % colors
\usepackage{xspace}
\usepackage{bbding}

\usepackage{graphicx}
\usepackage{subfigure}

\usepackage{amsthm}
\usepackage{cite}
\usepackage{amsmath,amssymb,amsfonts}
\usepackage{algorithmic}
\usepackage{tabularx}

\usepackage{amsmath}
\usepackage{amssymb}
\usepackage{amsmath,bm}
\usepackage{color}
\usepackage{xcolor,soul}
\usepackage{algorithm}
\usepackage{multirow}
\usepackage{xspace}
\usepackage{soul}
\usepackage{wrapfig}

\newdimen\abovecrulesep
\newdimen\belowcrulesep
\makeatletter
\patchcmd{\@@@cmidrule}{\aboverulesep}{\abovecrulesep}{}{}
\patchcmd{\@xcmidrule}{\belowrulesep}{\belowcrulesep}{}{}

\definecolor{demphcolor}{RGB}{144, 144, 144}
\definecolor{mygray}{gray}{0.4}
\definecolor{lightgray}{rgb}{0.9, 0.9, 0.9}

\newlength\savewidth

\newcommand{\tablestyle}[2]{\setlength{\tabcolsep}{#1}\renewcommand{\arraystretch}{#2}\centering\small}
\makeatletter\renewcommand\paragraph{\@startsection{paragraph}{4}{\z@}{.5em\@plus1ex\@minus.2ex}{-.5em}{\normalfont\normalsize\bfseries}}
\makeatother

\usepackage{cleveref}
\crefname{section}{Sec.}{Secs.}
\Crefname{section}{Section}{Sections}
\Crefname{table}{Table}{Tables}
\crefname{table}{Tab.}{Tabs.}
\crefname{figure}{Fig.}{Figs.}
\Crefname{figure}{Figure}{Figures}

\usepackage{booktabs}
\usepackage{multirow}
\usepackage{tabularx}

\title{Youku-mPLUG: A 10 Million Large-scale Chinese Video-Language Pre-training Dataset and Benchmarks}

\newcommand{\modelname}{mPLUG-video\xspace}
\newcommand{\datasetname}{Youku-mPLUG\xspace}

\author{%
  Haiyang Xu\thanks{Equal contribution}, Qinghao Ye*, Xuan Wu*, Ming Yan\thanks{Corresponding author}, Yuan Miao, Jiabo Ye
  \\\textbf{Guohai Xu, Anwen Hu, Yaya Shi, Guangwei Xu, Chenliang Li, Qi Qian}  
  \\\textbf{Maofei Que, Ji Zhang, Xiao Zeng, Fei Huang} \\
 DAMO Academy, Alibaba Group \\
 \texttt{\{shuofeng.xhy, yeqinghao.yqh, wx193834, ym119608\}@alibaba-inc.com} \\
}

\begin{document}

\maketitle

\begin{abstract}
To promote the development of Vision-Language Pre-training (VLP) and multimodal Large Language Model (LLM)  in the Chinese community, we firstly release the largest public Chinese high-quality video-language dataset named \datasetname, which is collected from Youku\footnote{\href{https://www.youku.com}{https://www.youku.com}}, a well-known Chinese video-sharing website, with strict criteria of safety, diversity, and quality. \datasetname contains 10 million Chinese video-text pairs filtered from 400 million raw videos across a wide range of 45 diverse categories for large-scale pre-training. In addition, to facilitate a comprehensive evaluation of video-language models, we carefully build the largest human-annotated Chinese benchmarks covering three popular video-language tasks of cross-modal retrieval, video captioning, and video category classification. \datasetname can enable researchers to conduct more in-depth multimodal research and develop better applications in the future. Furthermore, we release popular video-language pre-training models, ALPRO and mPLUG-2, and our proposed modularized decoder-only model \modelname pre-trained on \datasetname. Experiments show that models pre-trained on \datasetname gain up to 23.1\% improvement in video category classification.%significant improvement. 
~
Besides, \modelname achieves a new state-of-the-art result on these benchmarks with 80.5\% top-1 accuracy in video category classification and 68.9 CIDEr score in video captioning, respectively. Finally, we scale up \modelname based on the frozen Bloomz with only 1.7\% trainable parameters as Chinese multimodal LLM, and demonstrate impressive instruction and video understanding ability. The zero-shot instruction understanding experiment indicates that pretraining with \datasetname can enhance the ability to comprehend overall and detailed visual semantics, recognize scene text, and leverage open-domain knowledge.
Our dataset, code, model, and evaluation set are available at \href{https://github.com/X-PLUG/Youku-mPLUG}{https://github.com/X-PLUG/Youku-mPLUG}. 
  
\end{abstract}

\vspace{-3ex}
\section{Introduction}
Due to the release of large-scale English video-language datasets (e.g., Howto100M\citep{howto100m} and WebVid-2.5M\citep{bain2021frozen}), video-language pre-training (VLP) has achieved superior performance in various downstream tasks, such as video-text retrieval, video question answering, and video captioning. Recently, the multimodal LLM in video (e.g., VideoChat\citep{li2023videochat}, Flamingo\citep{alayrac2022flamingo}) has demonstrated strong zero-shot video understanding ability based on these large-scale datasets. Compared with the English VLP community as \cref{table:pretrain_compare}, the lack of large-scale and high-quality public Chinese VLP datasets hinders the research of Chinese video-language pretraining and multimodal LLM. In addition, the Chinese VLP community is also facing a lack of publicly available benchmarks as \cref{table:downstream_compare}. This has resulted in two significant issues. Firstly, the development and application of the community are being lagged behind. Secondly, some works are able to achieve surprisingly good performance by using secret downstream benchmarks that other works cannot fairly compare with, thus making it difficult to establish standards for performance evaluation. While some methods translate English text into Chinese \citep{Madasu2022ImprovingVR} or annotate the dataset based on the English video \citep{xinwang2019vatex}, there remains an intrinsic linguistic and cultural gap between English and Chinese. 

\begin{table}[]
\centering
\caption{Statistics of \datasetname and its comparison with existing video-language pre-training datasets.}
    \tablestyle{7pt}{1.1} 
    \def \w{15pt}
    \resizebox{\textwidth}{!}{
        \begin{tabular}{l|cccccc|c} \hline
        \textbf{Dataset Name}                & \textbf{Language} & \textbf{\# Videos} & \textbf{\# Text} & \textbf{Avg. Len (secs)} & \textbf{Duration (hrs)} & \textbf{Domain}      & \textbf{Availability} \\ \hline
        HowTo100M \citep{howto100m}                & English  & 136M      & 136M    & 3.6             & 135K           & Instruction & \Checkmark \\
        YT-Temporal-180M \citep{zellers2021merlot}        & English  & 180M      & 180M    & -               & -              & Instruction & \Checkmark \\
        HD-VILA-100M \citep{xue2022hdvila}            & English  & 103M      & 103M    & 13.4            & 372K           & Open        & \Checkmark \\
        WebVid10M \citep{bain2021frozen}                & English  & 10M       & 10M     & 18.0            & 52K            & Open        & \Checkmark \\ \hline
        ALIVOL-10M \citep{Chenyi2021ALIVOL}               & Chinese  & 103M      & 110M    & 34.6             & 99K            & E-Commerce  &  \XSolidBrush     \\
        Kwai-SVC-11M \citep{Liqiang2022KwaiSVC}               & Chinese  & 11M       & 4M      & 57.9             & 177K           & Open        & \XSolidBrush     \\
        CREATE-10M \citep{zhang2022create}                 & Chinese  & 10M       & 10M     & 29.8             & 83K            & Open        &   \XSolidBrush   \\
        CNVid-3.5M \citep{Gan_2023_CVPR}                 & Chinese  & 3.5M      & 3.5M    & 36.2            & 35K            & Open        & \XSolidBrush     \\ \hline \hline
        \textbf{\datasetname} & Chinese  & 10M       & 10M     &  54.2             &  150K           & Open        & \Checkmark \\ \hline
        \end{tabular}
    }
    \label{table:pretrain_compare}
\end{table}

\begin{table}[]
\centering
\caption{Statistics of \datasetname and its comparison with existing video-language downstream datasets.}
    \tablestyle{7pt}{1.1} 
    \def \w{15pt}
    \resizebox{\textwidth}{!}{
        \begin{tabular}{l|ccc|ccc|c} \hline
        \textbf{Dataset Name} & \textbf{Language} & \textbf{\# Sample} & \textbf{Domain} & \textbf{Retrieval} & \textbf{Classification} & \textbf{Caption} & \textbf{Availability} \\ \hline
        MSRVTT \citep{xu2016msrvtt}                & English           & 10K               & Open            & \Checkmark         & \Checkmark              & \Checkmark       & \Checkmark            \\
        DiDeMo \citep{anne2017didemo}            & English           & 27K               & Flickr          & \Checkmark         & \XSolidBrush            & \XSolidBrush     & \XSolidBrush          \\
        MSVD \citep{chen2011msvd}            & English           & 10K               & Open            & \Checkmark         & \Checkmark              & \Checkmark       & \Checkmark            \\
        LSMDC \citep{rohrbach2015lsmdc}          & English           & 118K              & Movie           & \Checkmark         & \Checkmark              & \XSolidBrush     & \Checkmark            \\
        ActivityNet \citep{krishna2017ActivitynetDense}       & English           & 100K              & Open            & \Checkmark         & \Checkmark              & \XSolidBrush     & \Checkmark            \\ \hline
        VATEX \citep{xinwang2019vatex}           & English/Chinese   & 41K               & Kinetics-600    & \Checkmark       & \XSolidBrush            & \Checkmark       & \Checkmark          \\
        BFVD \citep{Shengyu2020Poet}           & Chinese           & 43K               & E-Commerce      & \Checkmark         & \XSolidBrush            & \XSolidBrush     & \XSolidBrush          \\
        FFVD \citep{Shengyu2020Poet}            & Chinese           & 32K               & E-Commerce      & \Checkmark         & \XSolidBrush            & \XSolidBrush     & \XSolidBrush          \\
        CREATE-210K \citep{zhang2022create}      & Chinese           & 216K              & Open            & \Checkmark         & \XSolidBrush            & \Checkmark       & \XSolidBrush          \\ \hline \hline
        \textbf{\datasetname}          & Chinese           & 365K              & Open            & \Checkmark         & \Checkmark              & \Checkmark       & \Checkmark          \\ \hline
        \end{tabular}
    }
    \label{table:downstream_compare}
\end{table}

To facilitate the research and application of Chinese VLP, we release the first and largest public Chinese Video-language pretraining dataset and benchmarks named \datasetname, which is collected from Youku, a well-known Chinese video-sharing website. \datasetname contains 10 million video-text pairs for pre-training and 0.3 million videos for downstream benchmarks. For the pre-training dataset, we filter the high-quality 10 million video-text pairs from 400 million raw videos with the strict criteria of safety, diversity, and quality. \textbf{Safety}, the dataset is subject to heavy filtering and restrictions through a multi-level risk detection system to prevent any content related to high risk; \textbf{Diversity}, the videos are carefully classified into 45 diverse  categories covering various domains, with a balanced distribution; \textbf{Quality}, we have conducted strict data cleaning at both the text and video levels, and using Chinese image-text pre-trained model to improve the data quality. Furthermore, We build the largest human-annotated Chinese benchmarks covering Cross-modal Retrieval, Video Captioning and Video Category Classification for comprehensive evaluation of video-language models and downstream applications. For each downstream task, we hire well-educated people and adopt a two-step verification for ensuring the quality and diversity of the annotations, resulting in the largest Chinese downstream benchmark for model evaluation. 

Besides, we release popular video-language models, ALPRO\citep{li2022alpro} and mPLUG-2\citep{xu2023mplug2} pre-trained on \datasetname. Drawing inspiration
from the idea of modularization\citep{li2022mplug, xu2023mplug2, ye2023mplugowl}, we propose the modularized decoder-only model \modelname with limited trainable parameters based on frozen pre-trained LLM, which consists of the trainable video encoder, visual abstractor module, and the frozen pre-trained LLM decoder. We first obtain dense video representations from the video encoder. Then, we employ the visual abstractor module to summarize visual information within several learnable tokens. Finally, the visual representations are combined with text queries and fed into the LLM decoder to generate the response. Experiments show that models pre-trained on \datasetname gain up to 23.1\% improvement in video category classification. Besides, \modelname achieves 80.5\% top-1 accuracy in video category classification and 68.9 CIDEr score in video captioning, respectively. It becomes the new state-of-the-art results on these benchmarks. Moreover, we scale up \modelname based on frozen Bloomz\citep{workshop2023bloom} as Chinese multimodal LLM with only 1.7\% trainable parameters, which demonstrates impressive instruction and video understanding ability. 
As an insight, our zero-short video instruction understanding test validates that \datasetname can strengthen the scene text recognizing ability and incorporate open-domain knowledge for video understanding. Qualitative results can be found in the Supplementary Material.
These pre-trained models has also been released to facilitate the research and application of Chinese Video-language pre-training.

In summary, our main contributions are:
\begin{itemize}
    \item We release the first public largest Chinese Video-language pretraining dataset and benchmarks named \datasetname.
    \item We release popular video-language models (ALPRO and mPLUG-2) and our proposed modularized decoder-only model \modelname pre-trained on \datasetname.
    \item We scale up and release \modelname based on Bloomz as Chinese multimodal LLM  with only 1.7\% trainable parameters, which demonstrates the impressive zero-shot instruction and video understanding ability.
    \item Experiments show that models pre-trained on \datasetname gain significant improvement and \modelname achieves the new state-of-the-art results on these benchmarks.
\end{itemize}

% For text, we have imposed language restrictions on video titles, requiring the length to be between 5 and 30 words and include at least 5 Chinese characters, while filtering out those with obvious advertising or meaningless content. In terms of video quality and integrity, we have specifically chosen recently uploaded videos with durations ranging from 10 to 120 seconds to ensure clear and complete content.

% To guarantee the
% diversity and generalization, our \datasetname dataset is collected according to a high-frequency Chinese word list with 200K queries. We also adopt image-based and text-based filtering strategies for further refinement. The resulting dataset is currently the largest Chinese vision-language dataset. We perform an analysis of this dataset and show that it covers a wide range of visual and textual concepts. Furthermore, 

\vspace{-2ex}
\section{Related Work}
\vspace{-1ex}
\paragraph{Video-Language Pre-training Datasets}
% Large-scale datasets have been shown effective for video-language representation learning. In the early stage, most video-language models are trained on HowTo100M dataset \citep{} which consists of 136 million video clips from 1.22 million instructional YouTube videos. However, the domain of this dataset is limited to instructional videos, which is not suitable for generalization. To overcome the constraint of the instructional domain, Zeller et al. \citep{} and Xue et al. \citep{} propose YT-Temporal-180M and HD-VILA-100M corpus respectively. Meanwhile, to reduce the noise in the subtitles, Bain et al. \citep{} introduce Webvid10M inspired by the collection schemes of Conceptual Caption datasets \citep{}. However, these datasets only contain English corpus which cannot be directly applied to the Chinese domain. Although there are some large-scale Chinese video-language datasets such as ALIVOL \citep{}, Kwai-SVC \citep{}, CREATE10M \citep{}, and CNVid-3.5M \citep{}, none of them are publicly released until now, which hinders the progress of research in the Chinese video-language learning field. To this end, we release the largest Chinese high-quality video-language dataset named \datasetname to facilitate future research on large-scale video-language learning in Chinese.
Large-scale datasets have proven effective for video-language representation learning. Previously, most video-language models were trained on the HowTo100M dataset \citep{howto100m}, which comprises 136 million video clips from 1.22 million instructional YouTube videos. However, this dataset is limited to the instructional domain and is unsuitable for generalization. To overcome this constraint, Zeller et al. \citep{zellers2021merlot} and Xue et al. \citep{xue2022hdvila} propose the YT-Temporal-180M and HD-VILA-100M corpus, respectively. Meanwhile, to reduce the noise in subtitles, Bain et al. \citep{bain2021frozen} introduce the Webvid10M dataset which is inspired by the collection schemes of Conceptual Caption datasets \citep{sharma-etal-2018-conceptual}. However, these datasets are limited to English language corpus and cannot be directly applied to the Chinese domain. Although there exist some large-scale Chinese video-language datasets such as ALIVOL \citep{Chenyi2021ALIVOL}, Kwai-SVC \citep{Nie2022SearchorientedMC}, CREATE-10M \citep{zhang2022create}, and CNVid-3.5M \citep{Gan_2023_CVPR}, none of them have been publicly released to date, which hinders the progress of research in the Chinese video-language learning field. To address this gap, we present \datasetname, the largest Chinese high-quality video-language dataset, to facilitate future research on large-scale video-language learning in the Chinese language.

\vspace{-2ex}
\paragraph{Video-Language Downstream Benchmarks}
For evaluating video-language pre-training models, researchers have proposed several downstream tasks such as video-text retrieval, video question answering, and video captioning for performance evaluation. For instance, MSRVTT \citep{xu2016msrvtt}, DiDeMo \citep{anne2017didemo}, and LSMDC \citep{rohrbach2015lsmdc} are commonly adopted for text-video retrieval evaluation. Similarly, MSRVTT-QA \citep{xu2017msrvttqa}, MSVD-QA \citep{xu2017msrvttqa}, and T-GIF \citep{jang2017tgif} are widely used for video question evaluation. Meanwhile, MSRVTT-Caption \citep{xu2016msrvtt} and MSVD-Caption \citep{chen2011msvd} are commonly used for video caption evaluation. However, these datasets are primarily collected from YouTube, which is not entirely suitable for the Chinese domain. Furthermore, while there are some Chinese benchmark datasets such as CREATE \citep{zhang2022create} and VATEX \citep{xinwang2019vatex}, they are not fully released and only evaluate one aspect of the model's performance. Additionally, there is a lack of systematic video language downstream benchmarks or leaderboards for Chinese video-language pre-training evaluation. Consequently, we propose three downstream benchmarks, including video category classification, video-text retrieval, and video captioning, for evaluating models' performance on \datasetname. These benchmarks are specifically designed for the Chinese domain and are intended to fill the gap in existing English benchmarks, which may not be entirely suitable for Chinese video-language pre-training evaluation.

\vspace{-3ex}
\paragraph{Video-Language Pre-training Models}
In recent years, there has been a growing interest in video-language pre-training, and various methods have been proposed to explore this area. Traditional approaches \citep{Luo2020UniViLMAU, Li2020HeroHE} rely on pre-extracted, dense video frame or clip features for video-language representation. In contrast, ClipBERT \citep{lei2021clipbert} introduces a sparse sampling strategy that facilitates end-to-end learning while simultaneously improving performance. Building upon this strategy, many approaches \citep{bain2021frozen, ge2022bridgeformer} have been developed, which incorporate novel architectures and pre-training tasks for video-language learning. For example, Frozen \citep{bain2021frozen} and BridgeFormer \citep{ge2022bridgeformer} employ contrastive learning to align the semantics of paired video and text in the same embedding space. Additionally, ALPRO \citep{li2022alpro}, TW-BERT \citep{yang2023twbert}, mPLUG-2 \citep{xu2023mplug2}, and HiTeA \citep{ye2022hitea} fuse video and language features to generate video-language representations for understanding and generation. Recently, large language models such as GPT-3 \citep{brown2020gpt3}, Bloom \citep{workshop2023bloom}, and LLaMA \citep{touvron2023llama} have demonstrated significant zero-shot generalization abilities, which are advantageous for the vision-language field. For instance, BLIP-2 \citep{li2023blip2}, miniGPT-4 \citep{zhu2023minigpt}, and mPLUG-Owl \citep{ye2023mplugowl} exhibit robust zero-shot generalization and conversation capabilities by aligning vision and language models. In this work, we provide a decoder-only video-language model \modelname pre-trained on our \datasetname dataset with a strong generalization performance in terms of both video-language understanding and generation.

\vspace{-2ex}
\section{Youku-mPLUG Dataset Creation}
\begin{figure}[!htp]
    \centering
    \includegraphics[width=0.8\linewidth]{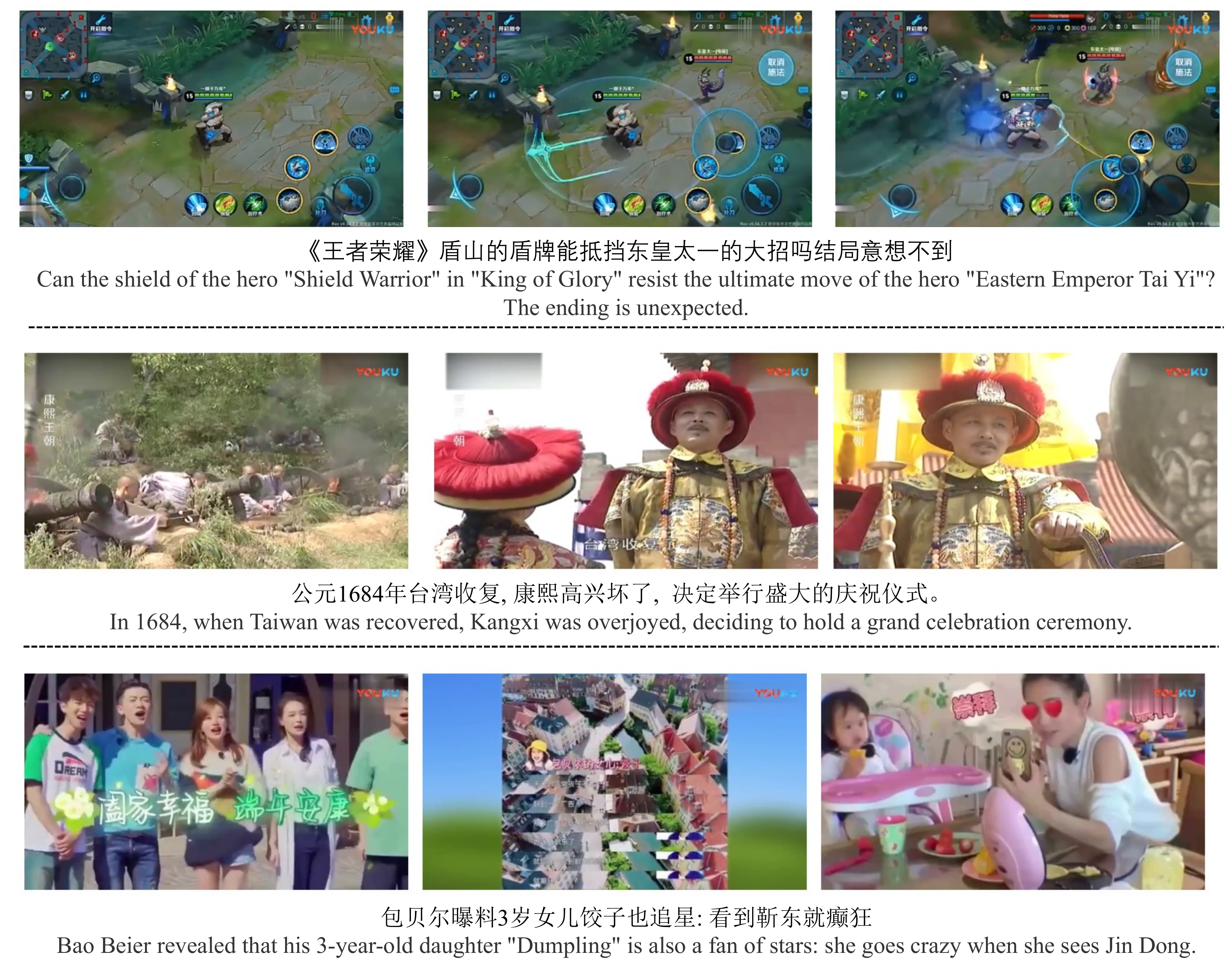}
    \vspace{-2ex}
    \caption{Random sampled examples in \datasetname.}
    \label{fig:youku_case}
\end{figure}

\vspace{-2ex}
To fill in the blank of the public Chinese video-text pre-training dataset and benchmarks, We release the largest public Chinese Video-language dataset named \datasetname collected with the strict criteria of safety, diversity, and quality from Youku, a Chinese video-sharing website. \datasetname contains 10 million video-text pairs for pre-training and 0.3 millon videos for downstream benchmarks covering Video-Text Retrieval, Video Captioning and Video Category Classification. Randomly sampled examples are shown in \Cref{fig:youku_case}.

\begin{figure}[!htp]
    \centering
    \includegraphics[width=\linewidth]{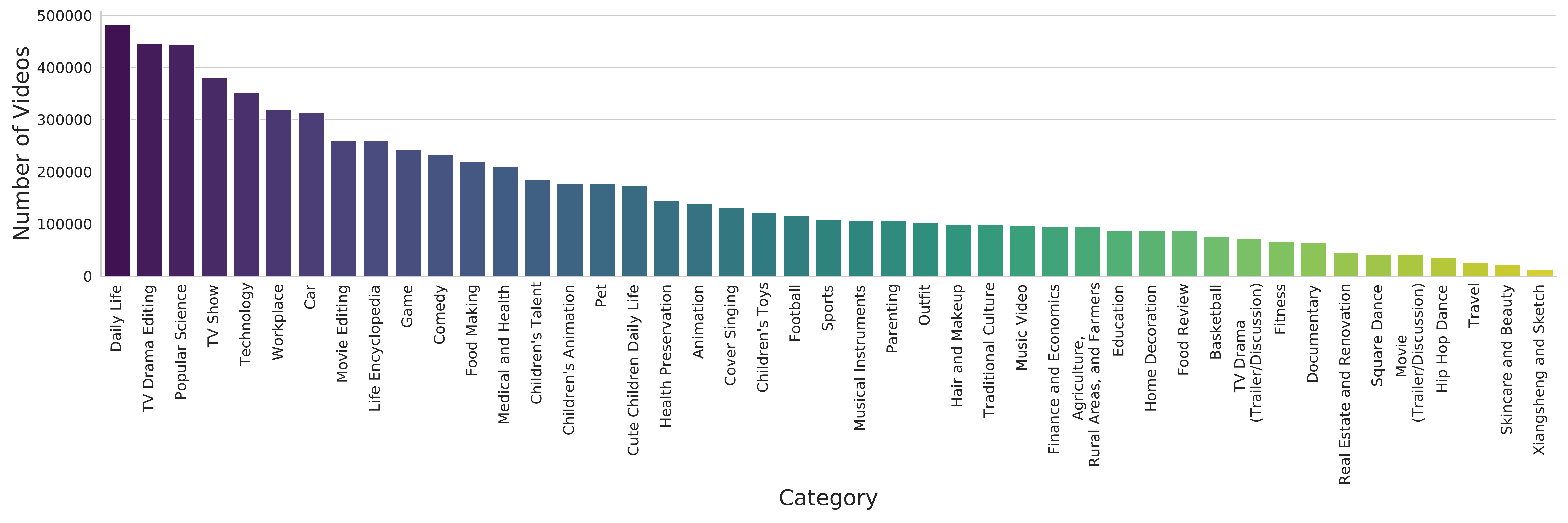}
    \vspace{-4ex}
    \caption{The distribution of the number of videos in each common category.}
    \label{fig:category-distribution}
\end{figure}

\begin{figure}[!htp]
    \begin{minipage}[c]{0.333\linewidth}
        \centering
        \includegraphics[width=\linewidth]{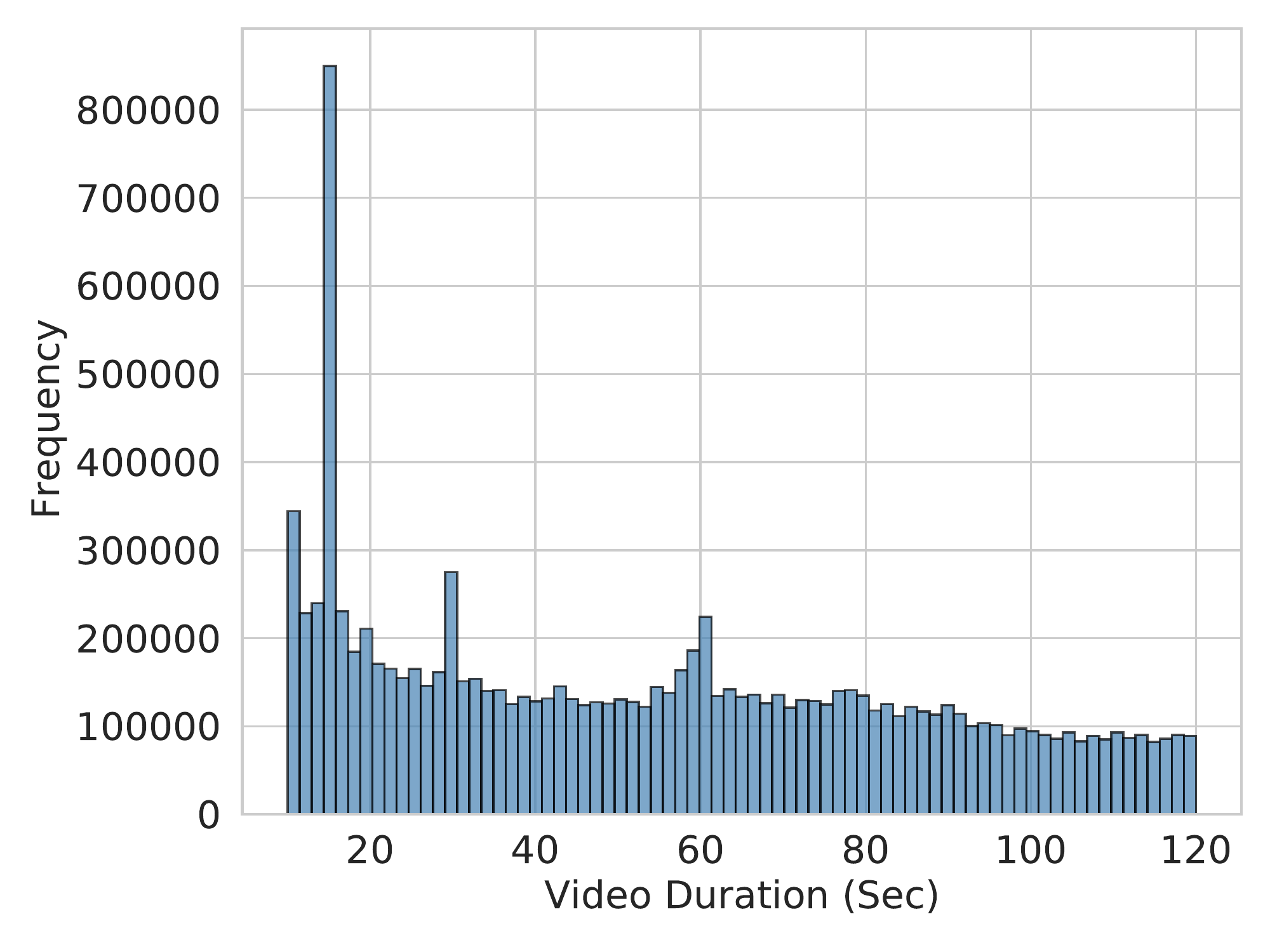}
        % \caption{XXXX}
    \end{minipage}\hfill
    \begin{minipage}[c]{0.333\linewidth}
        \centering
        \includegraphics[width=\linewidth]{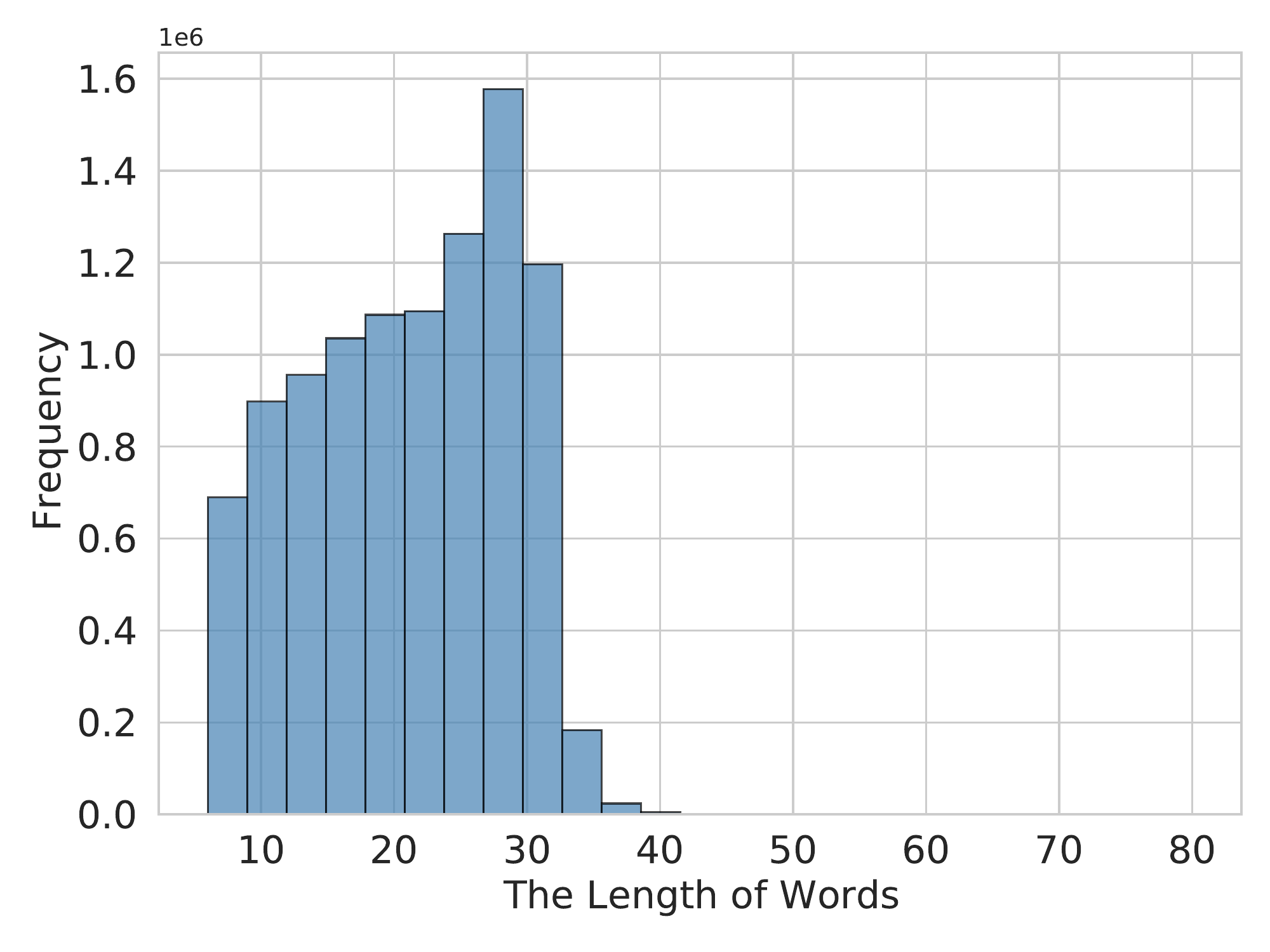}
        % \caption{XXXX}
    \end{minipage}\hfill
    \begin{minipage}[c]{0.333\linewidth}
        \centering
        \includegraphics[width=\linewidth]{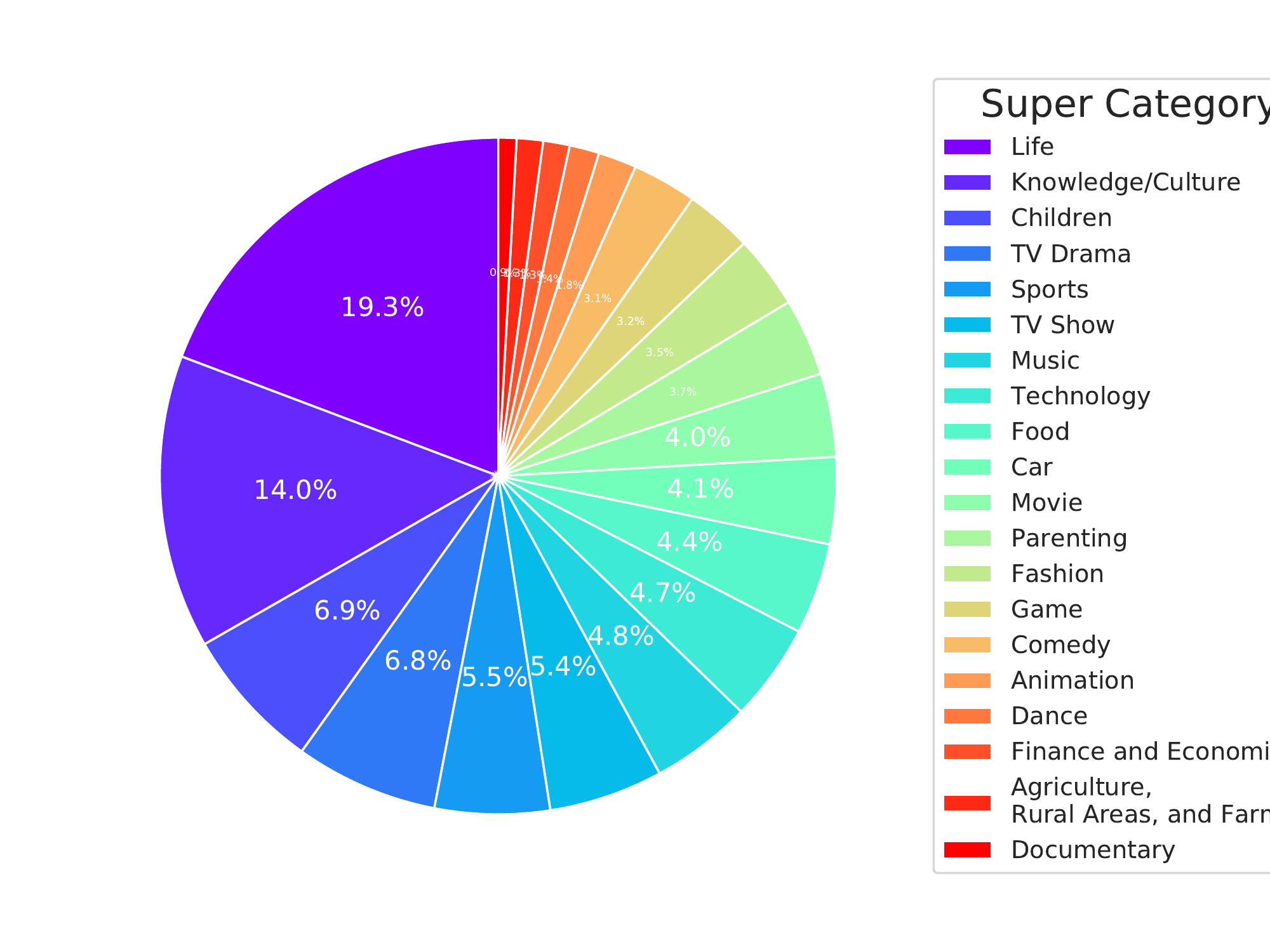}
        % \caption{XXXX}
    \end{minipage}\hfill
    \caption{\datasetname dataset statistics: we report the histogram of video duration in seconds (left), the histogram of
title length in words (middle), and the ratios of the categories in each super-category (right).}
 \label{fig:super-category}
\end{figure}

\vspace{-4ex}
\subsection{Pre-training Dataset Construction}
\vspace{-2ex}
For pre-training dataset, we filter the high-quality 10 million video-text pairs from 400 million raw videos with the strict criteria of safety, diversity, and quality.
In terms of safety, the dataset is heavily filtered and restricted by an internal multi-level risk detection system with both multimodal model detection and manual review processes to prevent any content related to pornography, violence, terrorism, discrimination, abuse, or high risk. Regarding diversity, we have applied video fingerprinting technology to eliminate videos that are completely identical. With the ability of hierarchical multi-label classification model \citep{giunchiglia2020neurips}, the videos are carefully classified into 20 super categories and 45 common categories as \cref{fig:category-distribution}, covering various domains, with a balanced distribution. To ensure high quality, we have conducted strict data cleaning at both the text and video levels. For text, we have imposed language restrictions on video titles, requiring the length to be between 5 and 30 words and include at least 5 Chinese characters while filtering out those with obvious advertising or meaningless content. In terms of video quality and integrity, we have specifically chosen recently uploaded videos with durations ranging from 10 to 120 seconds to ensure clear and complete content. Further, we also employ the Chinese image-text pre-trained model CLIP \citep{Yang2022ChineseCC} to improve the data quality by deprecating those with low similarities between the frame mean features and text features. \cref{fig:super-category} shows the statistics of video duration and word length.

\vspace{-2ex}
\subsection{Downstream Benchmark Construction}
\vspace{-2ex}
For the downstream benchmark, we design three types of tasks including video-text retrieval, video category classification, and video captioning to evaluate the models' performance in terms of understanding and generation. The statistics of these three different datasets are summarized in \cref{table:train_val_test}.

\begin{table}[!ht]
\centering
\caption{Statistics of \datasetname benchmark datasets. \# pairs indicates the number of video-text pairs.}
    \tablestyle{7pt}{1.1} 
    \def \w{15pt}
    % \resizebox{\textwidth}{!}{
        \begin{tabular}{l|c|c|c} \hline
\textbf{Task}                      & \textbf{Train (\# Pairs)} & \textbf{Val (\# Pairs)} & \textbf{Test (\# Pairs)} \\ \hline
Video Category Classification & 100,023          & 14,678         & 20,026          \\
Video-Text Retrieval      & 37,595           & 1,795          & 7,414           \\
Video Captioning          & 170,866          & 7,510          & 7,705          \\ \hline
\end{tabular}
    % }
    \label{table:train_val_test}
\end{table}

\paragraph{Video Category Classification} 
Our initial step involves randomly selecting a substantial number of videos based on category frequency. Next, we collect the video categories from the Youku database, which are auto-generated by an online model. It is important to note that this model's accuracy is approximately 94\% when considering historical prediction data, thus not entirely reliable. Consequently, we put forth additional efforts to ensure the quality and accuracy of our datasets by manually verifying each video and its corresponding title in the benchmark datasets. Prior to annotation, we supply a smaller dataset containing 100 videos, along with their metadata, including titles and categories generated by the online prediction model. Annotators are then tasked with confirming the assigned categories in relation to the videos and their titles. They must also assign a relevance score, which ranges from 1 to 5. A higher score suggests a greater likelihood of the video belonging to the given category, and those with scores above 3 are retained. Annotators with error rates exceeding 2.5\% are disqualified. After eliminating unsuitable annotators, we proceed with annotating the video category classification dataset. To ensure the utmost accuracy, particularly for the validation and testing sets, we engage three annotators to verify each video.

\vspace{-2ex}
\paragraph{Video Captioning}
The video captioning task requires the model to generate a concise sentence describing a video clip's content and title. To create the dataset, we randomly sample around 80,000 videos based on category frequency distribution and employ a color histogram-based approach for segmenting each video into shots \citep{Mei2014MultimediaSR}. To ensure an accurate understanding of the video content and produce precise descriptions, we engage several annotators who are native Chinese speakers with strong educational backgrounds. As part of the pre-annotation process during the video category classification task, we assign 25 random videos to each annotator, requesting them to create captions that include the subject and object in the video, as well as relevant descriptions of actions and background. The captions must consist of at least 15 Chinese characters. Following the pre-annotation stage, annotators proceed with annotating the datasets and split them into the training, validation, and testing sets. Especially, to prevent data leakage, clips from the same video or sharing the same title are exclusively assigned to either the training or testing sets. Moreover, for the validation and testing datasets, we enlist more than three individuals to annotate the video clips, promoting diversity and quality. 
% Furthermore, to guarantee the quality of Chinese captions, we implement a stringent two-stage verification process in which each collected description must be reviewed and approved by another independent worker. Workers with approval rates below 90\% are disqualified.

\vspace{-2ex}
\paragraph{Video-Text Retrieval} 
Given that we have already annotated video captions during the creation of the video captioning dataset, we select a subset of these annotated captions as text queries for video-text retrieval. Additionally, video titles can also be incorporated into the text queries, enhancing diversity within the text. In a similar vein, we ensure that clips from the same video or those with identical text titles are not exclusively included in the training or test set to prevent potential data leakage.

% To split the dataset into training, validation, and testing sets, we separate the video clips according to the corresponding searched queries. The clips from the same video or the same queries will not appear solely in the training or testing set to avoid overfitting.

\vspace{-3ex}
\section{Methodology}
\begin{figure*}[]
    \centering
    \includegraphics[width=0.9\textwidth]{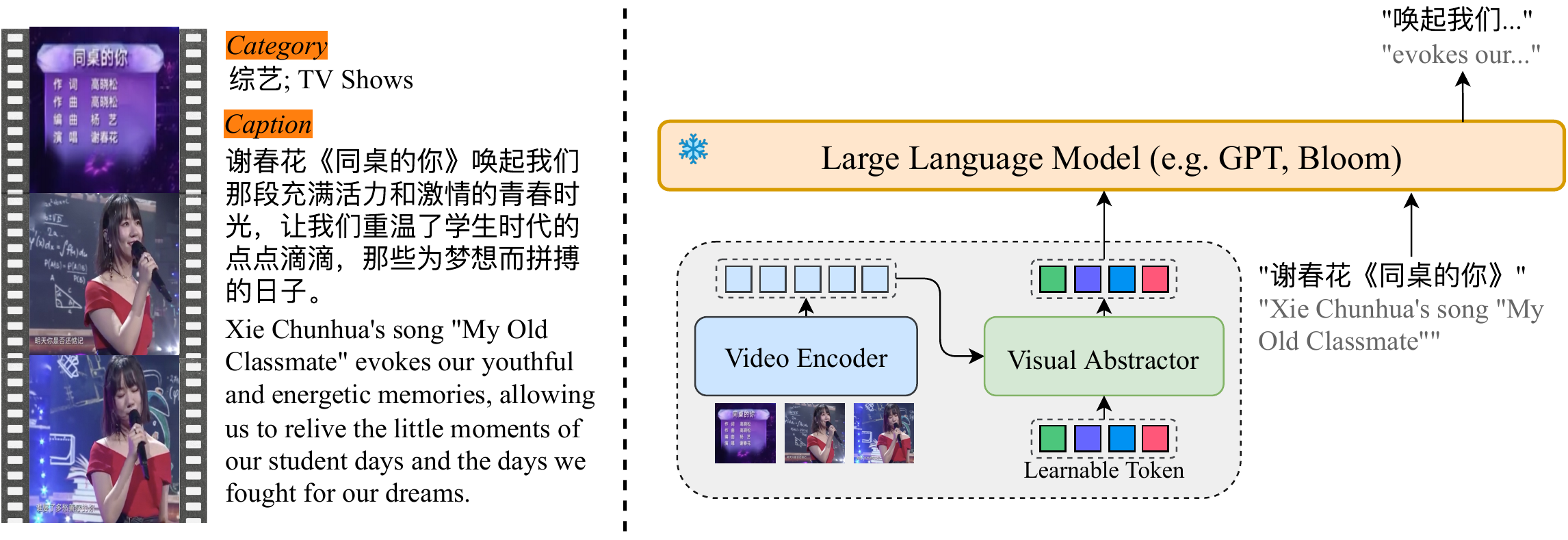}
    \caption{The overview of \modelname.}
    \label{fig:mPLUG-video}
    \vspace{-2mm}
\end{figure*}

\vspace{-2ex}
Since the pre-trained large language model shows incredible zero-shot and generalization abilities on various tasks, we use the off-the-shelf Chinese large language model (e.g. GPT-3 \citep{brown2020gpt3}) for efficient modularized training. To this end, we propose \modelname, a decoder-only based video-language model that leverages the frozen large language model. Especially, our model consists of a video encoder, a visual abstractor module, and a language decoder, as illustrated in Figure \ref{fig:mPLUG-video}. Besides, we leave the video encoder and visual abstractor trainable resulting in limited trainable parameters while largely reducing the computation burden.

\vspace{-2ex}
\subsection{Architecture}
\paragraph{The Video Encoder} We leverage a 12-layer TimeSformer \citep{bertasius2021timesformer} to extract the video features, with 224 height and width of each input frame. We sparsely sample $T$ frames from each video $\mathcal{V}$, the TimeSformer first divides the video frames into $N$ non-overlapping patches and flattens them into a sequence of $T\times N$ patches. Then these patches are fed into the patch projection layers for patch representation. To encode the position of each patch, we add learnable embeddings to encode each patch's spatial and temporal position. Then the TimeSformer applies divided spatiotemporal attention to yield video representation $V\in \mathbb{R}^{(T\times N) \times D}$, where $D$ is the hidden dimension of the video representation. 
\paragraph{Visual Abstractor Module} 
To mitigate the computation burden with the lengthy video sequences, we introduce visual abstractor module which utilizes learnable queries $Q \in \mathbb{R}^{M\times D}$ for reducing the length of video sequence as follows:
\begin{align}
    \Tilde{Q} &= CrossAttention(Q, V, V), \\
    \Tilde{Q} &= FFN(\Tilde{Q}) + \Tilde{Q},
\end{align}
where $CrossAttention(x, y, z)$ is the cross-attention layer with Query $x$, Key $y$, and Value $z$. The $FFN(\cdot)$ is the feed-forward layer \citep{Vaswani2017Attention}. Finally, we obtain the reduced video sequence $\Tilde{Q}\in \mathbb{R}^{M \times D}$.

\vspace{-2ex}
\paragraph{The Language Decoder} Since pre-trained large language models demonstrate strong zero-shot capabilities in text generation, we utilize them as the general text decoder for multi-modal inputs while keeping it frozen. In specific, we treat the video as a foreign language and concatenate the reduced video sequence with the text token features obtained from the text embedding layer. Then, the video and text token features are jointly fed into the large language model which is frozen for obtaining the video-guided language features. Finally, the video-guided language features are predicted for text tokens.

\vspace{-2ex}
\paragraph{Training Objective}
We train \modelname within an auto-regressive manner and adopt the next prediction task for training. In detail, the model needs to complete the texts based on the given video, and the language modeling loss is calculated as:
\begin{equation}
    \mathcal{L} = -\mathbb{E}_{(\mathcal{W}, \mathcal{V})} \left[ \sum_{l=1}^L \log p(w_l | \mathcal{W}_{[0, l)}, \mathcal{V}) \right],
\end{equation}
where $L$ denotes the total number of words in the text, and $\mathcal{W}$ denotes the word tokens.

\vspace{-1.5ex}
\subsection{Applying to Downstream Tasks}
\vspace{-1.5ex}
\paragraph{Video Captioning}
Video captioning is considered an auto-regressive task. During the process of fine-tuning a video captioning dataset, the training objective remains the same as that during pre-training. 

\vspace{-1.5ex}
\paragraph{Video Category Classification}
We treat video category classification as a video caption task. Annotated category names of videos are regarded as ground-truth captions. We evaluate the accuracy of predictions based on whether the predicted category name exactly matches the ground-truth.

\vspace{-1.5ex}
\paragraph{Video-Text Retrieval}
In contrast to mPLUG-2, which includes a contrastive head and a matching head for the retrieval task, our \modelname cannot be directly used for retrieval tasks. Therefore, we input video-text pairs into the model and extract the feature of the last token. We obtain the matching score by applying an extra linear layer to the feature of the last token.

\vspace{-2ex}
\section{Experiments}
\vspace{-1ex}
\subsection{Implementation Details}
\modelname leverage the pre-trained popular Chinese GPT-3 \footnote{https://modelscope.cn/models/damo/nlp\_gpt3\_text-generation\_1.3B/summary} \footnote{https://modelscope.cn/models/damo/nlp\_gpt3\_text-generation\_2.7B/summary} as the language decoder, and the video encoder is pre-trained on ImageNet \citep{ridnik2021imagenet}. During pre-training, we sparsely sample 8 frames from each video preserving their order in-between, and resize them to 224 × 224. We use a batch size of 512 and train \modelname for 10 epochs. We adopt the AdamW optimizer with $\beta=(0.9, 0.98)$, and set the learning rate and weight decay to 1e-4 and 1e-3 respectively. We warm up the training with 2000 warm-up steps then decay the learning rate with the cosine schedule. For downstream tasks, we use a batch size of 128 and train \modelname for 10 epochs with a learning rate 2e-5. 

\vspace{-2ex}
\subsection{Evaluation Results on Downstream tasks}
\vspace{-1ex}
In this study, we evaluate the performance of ALPRO, mPLUG-2, and \modelname on video category classification, video captioning, and video-text retrieval. 
\vspace{-1ex}
\paragraph{Evaluation on Video Category Classification}
We assess the performance of ALPRO, mPLUG-2, and mPLUG-Video on video category classification tasks. We measure the top-1 and top-5 accuracy of each model. For the generation models, a generated category name that is exactly the same as ground truth can be regarded as a correct prediction. The comparison results are shown in \Cref{tab:main_cat_cap}. Our results reveal that mPLUG-Video achieves the highest accuracy, with a top-1 accuracy of 80.57\% and a top-5 accuracy of 98.15\%. Interestingly, \modelname (2.7B) outperforms \modelname (1.3B), highlighting the importance of natural language understanding with a larger LLM decoder.
\vspace{-1ex}
\paragraph{Evaluation on Video Caption}
We present in \Cref{tab:main_cat_cap} the performance of models on Video Caption. ALPRO does not have a decoder module. Therefore, its performance was not reported. The performance of mPLUG-Video and mPLUG-2 are compared based on various metrics, including METEOR, ROUGE, CIDEr, and BLEU-4. It is found that \modelname (2.7B) achieves higher scores than mPLUG-Video (1.3B) across all four metrics. Additionally, \modelname obtains higher scores than mPLUG-2 on BLEU-4. These results suggest that pre-trained language models are essential and video captioning tasks based on our dataset are still challenging for existing methods.
\vspace{-1ex}
\paragraph{Evaluation on Video-Text Retrieval}
\Cref{tab:main_retrieval} presents the performance comparison between models on video retrieval task. We observe that mPLUG-2 outperforms ALPRO, possibly due to the incorporation of universal layers that remove modality differences and generate superior uni-modal representations. We also notice that \modelname perform poorly on video retrieval task. Freezing the language model can hinder \modelname to extract cross-modal features. These findings imply that our dataset accurately measures the video-language modeling capability.
\begin{table}[!htp]
\caption{Comparison results on \datasetname. Video category prediction and video captioning, respectively. For video category prediction, top-1 and top-5 accuracy are reported. For video captioning, we report BELU-4, METEOR, ROUGE and CIDEr. * denotes the language model is frozen.}
\resizebox*{\textwidth}{!}{%
\begin{tabular}{@{}ccccccc@{}}
\toprule
                       & \multicolumn{2}{c}{Video Category Prediction}                           & \multicolumn{4}{c}{Video Captioning}                                                                   \\
\multirow{2}{*}{Model} & \multirow{2}{*}{Top-1 Acc.(\%)} & \multirow{2}{*}{Top-5 Acc.(\%)} & \multirow{2}{*}{BLEU-4} & \multirow{2}{*}{METEOR} & \multirow{2}{*}{ROUGE} & \multirow{2}{*}{CIDEr} \\
                       &                                 &                                 &                         &                         &                        &                        \\ \midrule
ALPRO                  & 78.15                           & 95.15                           & -                       & -                       & -                      & -                      \\
mPLUG-2                & 77.79                           & 92.44                           & 43.7                    & \textbf{27.6}           & 52.9                   & 67.7                   \\
mPLUG-Video (1.3B)*    & 80.04                           & 98.06                           & 46.4                    & 26.5                    & 52.9                   & 67.7                   \\
mPLUG-Video (2.7B)*    & \textbf{80.57}                  & \textbf{98.15}                  & \textbf{47.1}           & 26.7                    & \textbf{53.3}          & \textbf{68.9}          \\ \bottomrule
\end{tabular}%
}
\label{tab:main_cat_cap}
\end{table}

\begin{table}[!htp]
\centering
\caption{Comparison results on \datasetname. Video retrieval. We evaluate models on video retrieval (V2T) and text retrieval (T2V). we report the average of R@1, R@5 and R@10. * denotes the language model is frozen.}
\begin{tabular}{@{}ccccccc@{}}
\toprule
                       & \multicolumn{6}{c}{Video Retrieval}                                                                 \\
\multirow{2}{*}{Model} & \multicolumn{3}{c}{V2T}                          & \multicolumn{3}{c}{T2V}                          \\
                       & R@1            & R@2            & R@10           & R@1            & R@5            & R@10           \\ \midrule
ALPRO                  & 27.00          & 53.33          & 64.09          & 26.63          & 53.20          & 64.43          \\
mPLUG-2                & \textbf{38.45} & \textbf{65.48} & \textbf{75.18} & \textbf{38.45} & \textbf{65.48} & \textbf{75.18} \\
mPLUG-Video (1.3B)*     & 7.01           & 20.33          & 29.67          & 7.01           & 20.33          & 29.67          \\
mPLUG-Video (2.7B)*     & 7.62           & 21.24          & 31.39          & 7.62           & 21.24          & 31.39          \\ \bottomrule
\end{tabular}

\vspace{-3ex}
\label{tab:main_retrieval}
\end{table}
\vspace{-1ex}
\subsection{Ablation Study on Modalities}
\vspace{-1ex}
% \usepackage{booktabs}
% TODO need to be filled
\begin{table}[!htp]
\caption{Comparison of different modalities and~\datasetname on category classification task.}
\tablestyle{7pt}{1.1} 
    \def \w{15pt}
    \resizebox{0.9\textwidth}{!}{
        \begin{tabular}{@{}ccccc@{}}
        \toprule
        Vision Modality & Language Modality & \datasetname Pre-Trained & Top-1 Acc.(\%) & Top-5 Acc.(\%) \\ \midrule
        \Checkmark      & \XSolidBrush      & \XSolidBrush          & 63.51          & 89.89          \\
        \XSolidBrush    & \Checkmark        & \XSolidBrush          & 59.31          & 86.31              \\
        \Checkmark      & \Checkmark        & \XSolidBrush            & 69.40          & 90.07          \\
        \Checkmark      & \Checkmark        & \Checkmark            & \textbf{78.15}          & \textbf{95.15}          \\ \bottomrule
        \end{tabular}
    }
    \vspace{-2ex}
%\caption{Performance of models trained with data consists of different modility on Category Prediction task. Vision Modality denotes model that only use video frames to predict the category. Language Modality denotes that the model only use the caption of video to predict the category. And AliceMind Pre-Trained denotes model pretrained with Youku-AliceMind dataset.}
\label{tab:modality_dataset}
\end{table}

In this section, we investigate the contributions of different modalities to video-language modeling by leveraging the category classification task on our \datasetname. \Cref{tab:modality_dataset} presents the performance of the baseline model (ALPRO) trained with data of different modalities. Vision Modality and Language Modality denote the model is trained with and utilizes this modality of data (video frames or video captions). And \datasetname Pre-Trained refers to the model that pre-trained on \datasetname before fine-tuning. The results show that the performance of the model trained with the visual modality is higher than that of the language modality. This suggests that high-level language modalities may lose fine-grained visual clues, leading to failure in retrieval. Additionally, we observe that the model trained with both vision and language modalities achieves higher performance than unimodal models. This observation demonstrates the importance of modality complementarity in video understanding. Pre-training the model with \datasetname leads to a significant improvement in model performance, emphasizing the importance of our \datasetname.

\subsection{Human Evaluation of Zero-shot Video Instruction Understanding}

\begin{wrapfigure}{r}{0.5\linewidth}
    \centering
    \setlength{\abovedisplayskip}{1pt} 
    \setlength{\belowdisplayskip}{1pt} 
    % \vspace{-12pt}
    \includegraphics[width=\linewidth]{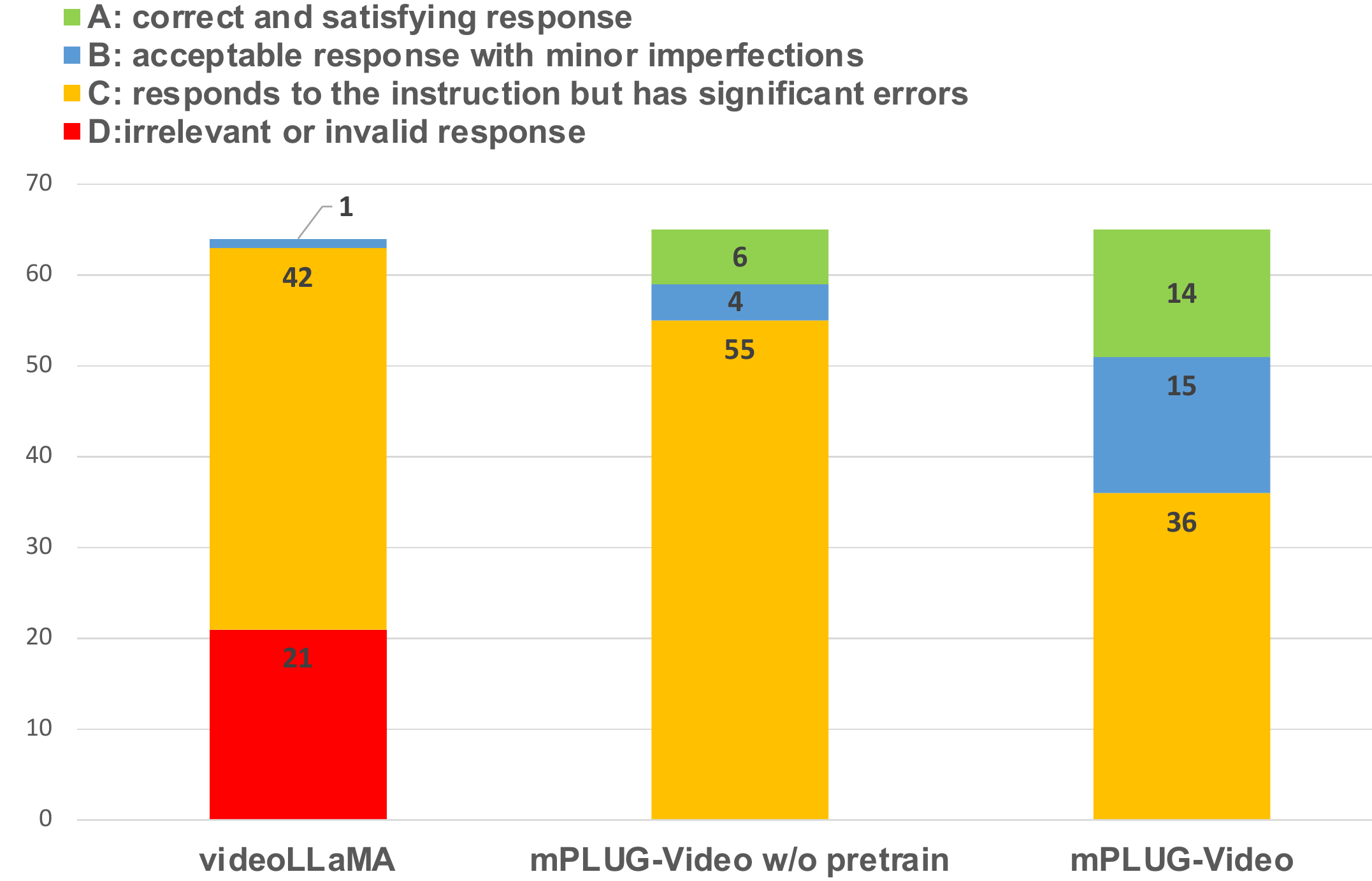}
    % \vspace{-12pt}
    \caption{Human evaluation about zero-shot video instruction understanding on 65 cases.}
    \label{fig: human_eval}
    % \vspace{-8pt}
\end{wrapfigure}

To test the video instruction understanding ability of different models, we manually set 65 instructions based on 50 randomly-sampled videos (45 from \datasetname, 5 from HD-VILA-100M \citep{xue2022hdvila}). 
%The instruction type covers video captioning, video question and answering. 
We compare the instruction understanding performance of three models: VideoLLaMA\citep{zhang2023videollama}, mPLUG-Video w/o pretrain and mPLUG-Video. VideoLLaMA is trained with visual instruction data from MiniGPT-4\citep{zhu2023minigpt}, LLaVa \citep{llava} and Video-Chat \citep{li2023videochat}, while the latter two models only utilize visual training data from LLaVa \citep{llava}. We ask human annotators to score the models' responses. Following Self-Instruct\citep{wang2022selfinstruct}, human annotators are asked to rate the response into four levels, where A means `correct and satisfying response', B means `acceptable response with minor imperfections', C means `response to the instruction but has significant errors' and D means `irrelevant or invalid response'. As shown in \cref{fig: human_eval}, with the pertaining in \datasetname, \modelname achieves much better video instruction understanding and responding ability, demonstrating the effectiveness of our proposed pretraining data. Qualitative results can be found in the supplementary material.

\vspace{-2ex}
% \subsection{Zero-shot Learning}
\section{Conclusion}
\vspace{-2ex}
%Video-Language Pre-training and Artificial Intelligence Generated Content have attracted attention of many researches. In this paper, we release the largest Chinese high-quality video-language dataset, namely Youku-AliceMind. We also public a human-annotated benchmark consists of three downstream tasks including Cross-modal Retrieval, Captioning, Category Classification. We propose a modularized decoder-only model mPLUG-video pre-trained on Youku-AliceMind. Experiments suggests that our evaluation set can effectively assess the video language comprehension and modeling ability of the models. Besides, pre-trained on Youku-AliceMind can gain significant improvements and our mPLUG-video achieves the new state-of-the-art.

In this paper, we introduce the largest high-quality video-language dataset in Chinese, called \datasetname. Additionally, we present a human-annotated benchmark that comprises three downstream tasks, i.e., Video-Text Retrieval, Video Captioning, and Video Category Classification. We propose a decoder-only model, \modelname, that is modularized and pre-trained on \datasetname. Results from our experiments indicate that our evaluation set can effectively evaluate the video language comprehension and modeling abilities of models. Furthermore, pre-training on \modelname leads to significant improvements, and our \modelname achieves a new state-of-the-art performance.

%To efficiently understand and utilize multi-media data, a amount of video-language models are proposed to better model video-language representation. 

\appendix
\section{Qualitative Results of Zero-shot Video Instruction Understanding}

\begin{figure}[!htp]
    \centering
    \includegraphics[width=\textwidth]{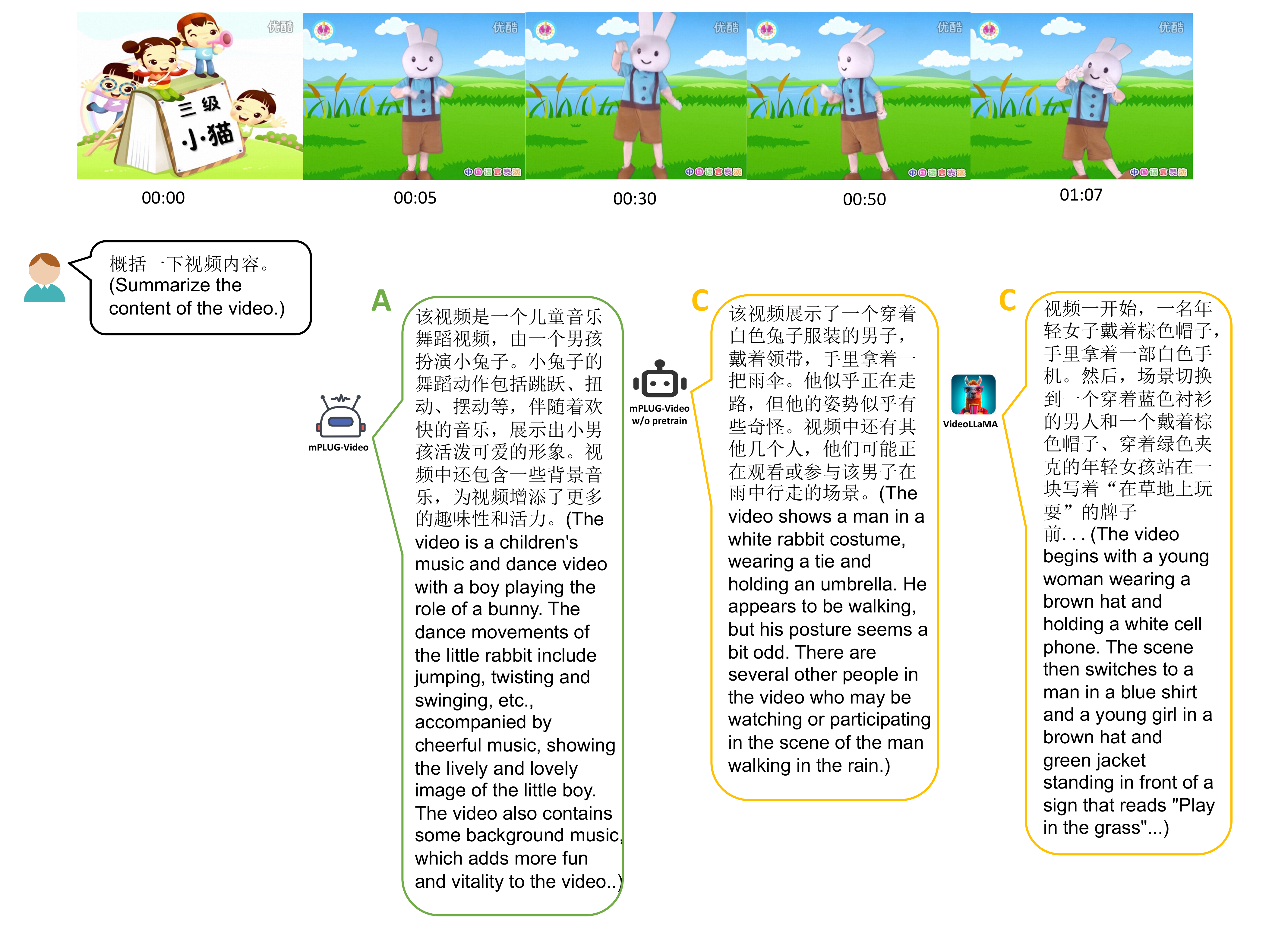}
    \caption{Qualitative results about video captioning.}
    \label{fig:zeroshot_cap}
    \vspace{-2mm}
\end{figure}

\begin{figure}[!tp]
    \centering
    \includegraphics[width=\textwidth]{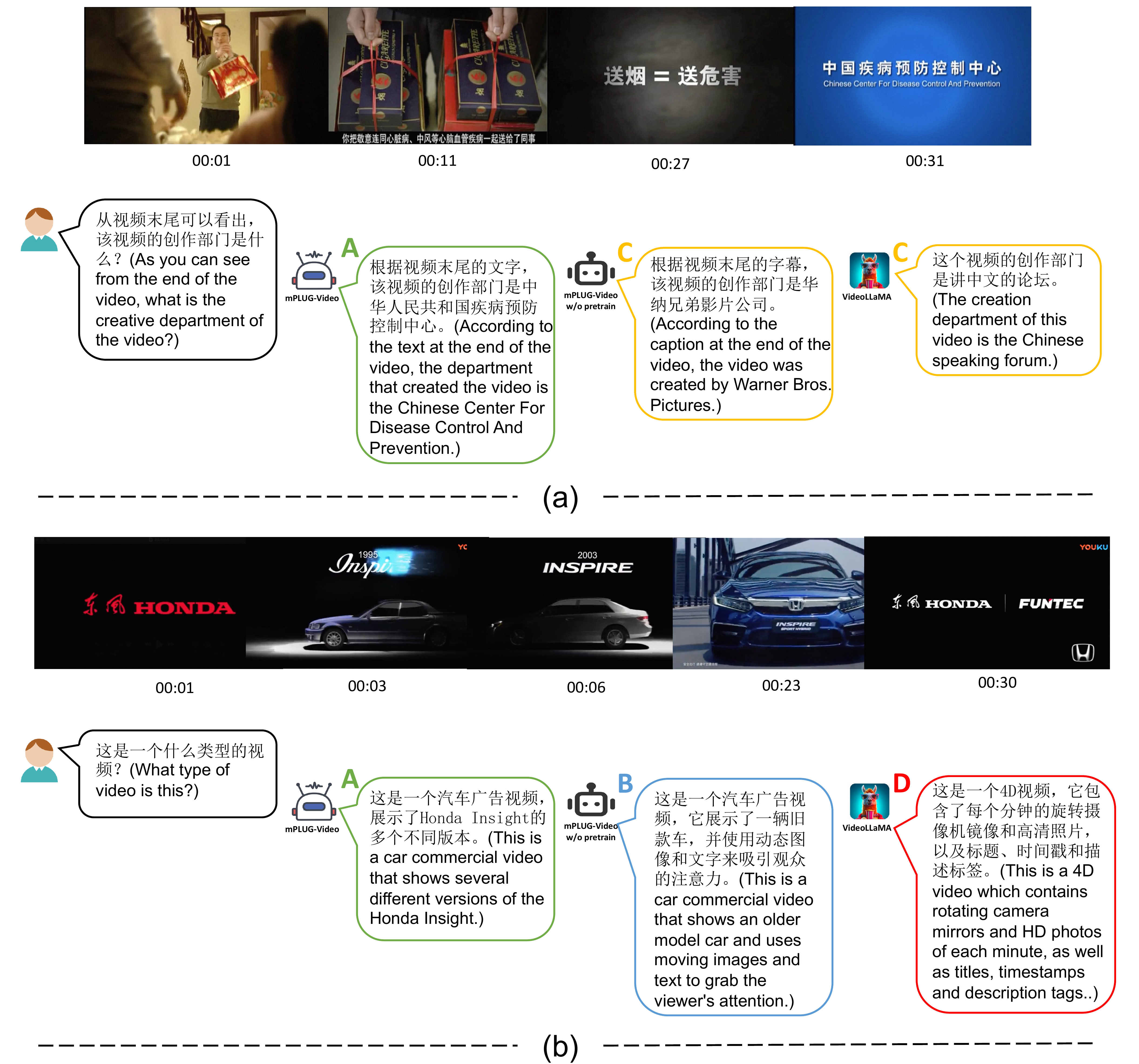}
    \caption{Qualitative results about video scene text understanding.}
    \label{fig:zeroshot_ocr}
    \vspace{-2mm}
\end{figure}

\begin{figure}[!htp]
    \centering
    \includegraphics[width=\textwidth]{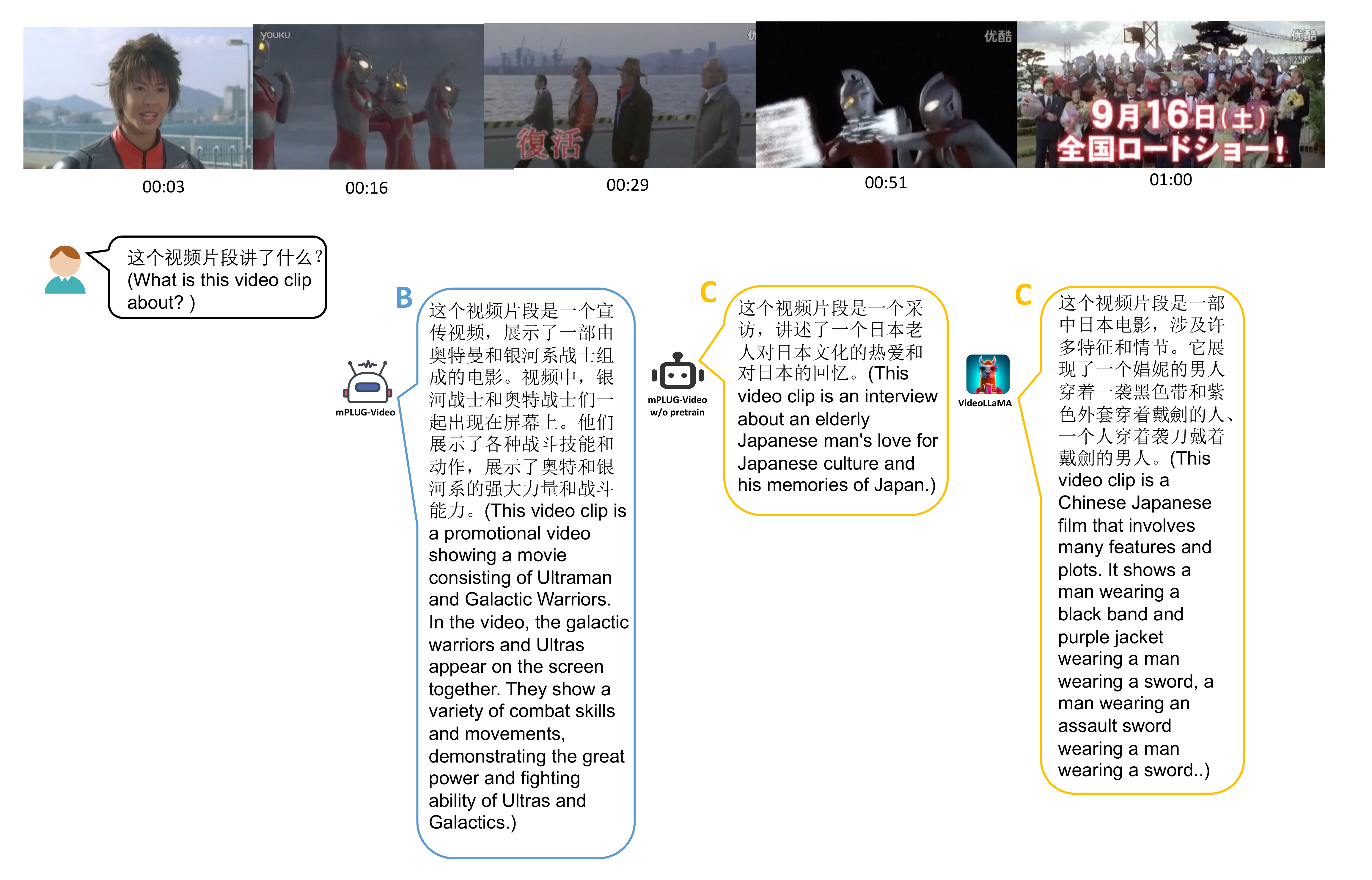}
    \caption{Qualitative results about open-domain knowledge understanding.}
    \label{fig:zeroshot_knowledge}
    \vspace{-2mm}
\end{figure}

In this section, we show some qualitative results of zero-shot video instruction understanding to validate the effectiveness of pretraining with \datasetname. Firstly, as shown in \cref{fig:zeroshot_cap}, with pertaining, \modelname can better understand both overall video semantic (`dance video') and detailed vision information, such as `jumping' and `twisting'. Besides, \cref{fig:zeroshot_ocr} shows that \modelname can better recognize scene texts in videos and give more accurate responses. Therefore, pretraining with \datasetname is highly likely to enhance the models' ability to comprehend scene text in videos. Furthermore, in \cref{fig:zeroshot_knowledge}, \modelname could accurately point out the key character name `Ultraman', while the other two models can't. This validates the contribution of pretraining for learning open-domain knowledge.

\begin{figure}[!htp]
    \centering
    \includegraphics[width=\textwidth]{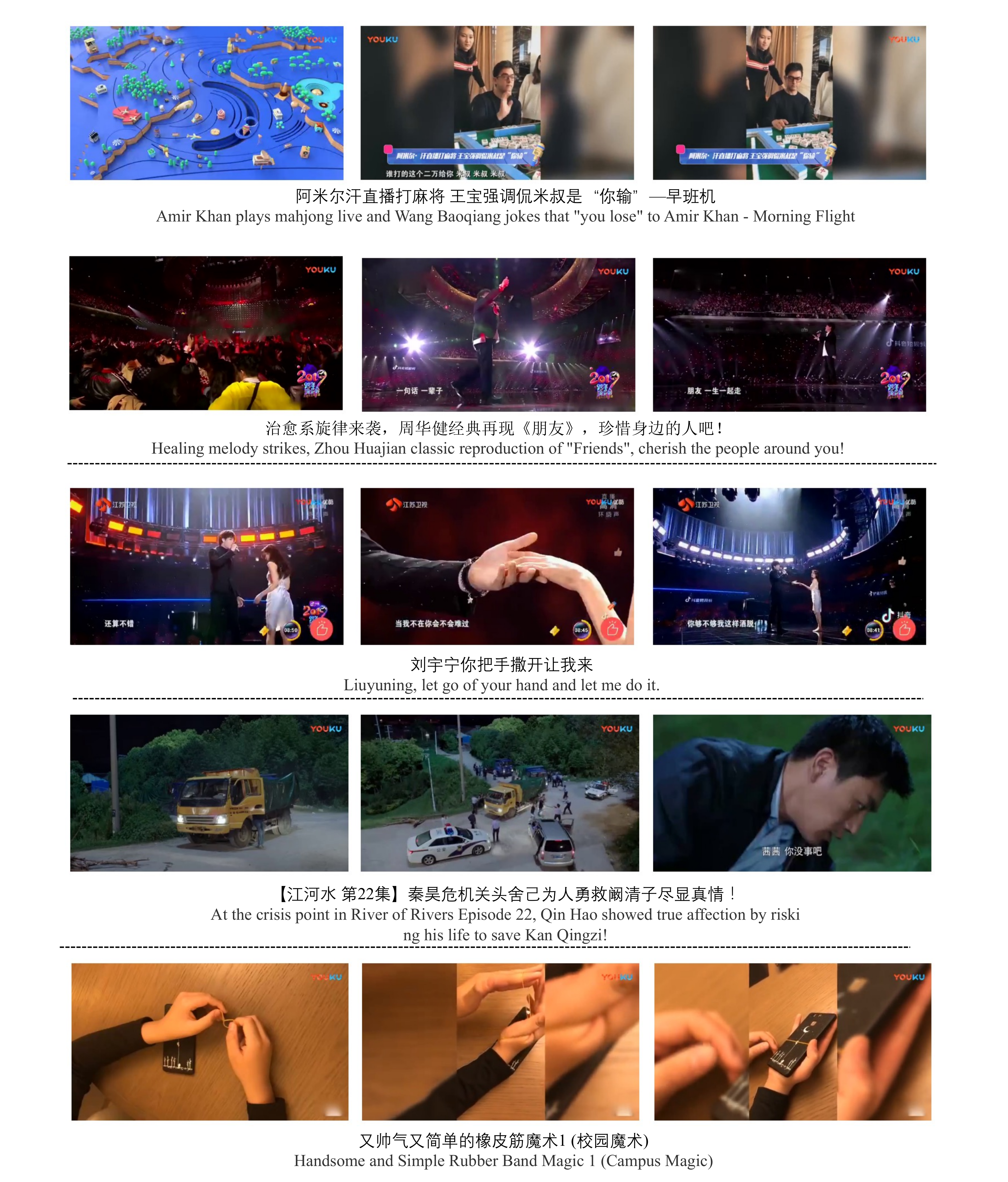}
    \caption{Examples in \datasetname.}
    \label{fig:example_full}
    \vspace{-2mm}
\end{figure}

\section{Limitations and Societal Impacts}
The \datasetname dataset predominantly contains concepts and language expressions that were current at the time of collection. As language evolves alongside human activities, our dataset may not encompass emerging concepts, words, and language expressions in the future. This limitation applies to image data as well, where new visual objects or designs might not be captured. Nevertheless, this issue can be addressed by fine-tuning pre-trained models on up-to-date data. Additionally, our dataset is constructed using corpora from the Chinese Internet, meaning the vocabulary and expressions may largely align with Chinese culture. Furthermore, our dataset lacks very long texts and lengthy videos, potentially limiting the ability of the pre-trained models to understand extensive content such as full-length movies.

\bibliographystyle{abbrvnat}
\clearpage
\bibliography{reference}
%%%%%%%%%%%%%%%%%%%%%%%%%%%%%%%%%%%%%%%%%%%%%%%%%%%%%%%%%%%%

\end{document}